\documentclass[format=sigconf]{acmart}
\AtBeginDocument{%
  }

 \setcopyright{none}

\usepackage{graphicx}  
\usepackage{lipsum}  
\usepackage{multirow} 
\usepackage{booktabs} 
\usepackage{tabularx} 
\copyrightyear{2025}
\acmYear{2025}
\setcopyright{rightsretained}
\acmConference[GECCO '25 Companion]{Genetic and Evolutionary Computation Conference}{July 14--18, 2025}{Malaga, Spain}
\acmBooktitle{Genetic and Evolutionary Computation Conference (GECCO '25 Companion), July 14--18, 2025, Malaga, Spain}\acmDOI{10.1145/3712255.3726658}
\acmISBN{979-8-4007-1464-1/2025/07}
\begin{document}

\title[Explainability of PSO Parameters]{Enhancing Explainability and Reliable Decision-Making in Particle Swarm Optimization through Communication Topologies}

\author{Nitin Gupta}

\orcid{0009-0000-2371-8363}
\affiliation{%
  \institution{Department of Mathematics and Computing\\ Dr BR Ambedkar NIT Jalandhar}
  \city{Jalandhar}
  \country{India}
}
\email{niting.ma.23@nitj.ac.in}
\author{Indu Bala}
\affiliation{%
  \institution{University of Adelaide}
  \city{Adelaide}
  \country{Australia}
}
\email{indu.bala@adelaide.edu.au}
\author{Bapi Dutta}
\affiliation{%
  \institution{University of Jaén}
  \city{Jaén}
  \country{Spain}}
\email{bdutta@ujaen.es}

\author{Luis Martínez}
\affiliation{%
  \institution{University of Jaén}
  \city{Jaén}
  \country{Spain}}\email{martin@ujaen.es}

\author{Anupam Yadav}
\authornotemark[1]
\affiliation{%
\institution{Department of Mathematics and Computing\\ Dr BR Ambedkar NIT Jalandhar}
  \city{Jalandhar}
  \country{India}}
  \email{anupam@nitj.ac.in}






\begin{abstract}
Swarm intelligence effectively optimizes complex systems across fields like engineering and healthcare, yet algorithm solutions often suffer from low reliability due to unclear configurations and hyperparameters. This study analyzes Particle Swarm Optimization (PSO), focusing on how different communication topologies—Ring, Star, and Von Neumann—affect convergence and search behaviors. Using an adapted IOHxplainer, an explainable benchmarking tool, we investigate how these topologies influence information flow, diversity, and convergence speed, clarifying the balance between exploration and exploitation. Through visualization and statistical analysis, the research enhances interpretability of PSO’s decisions and provides practical guidelines for choosing suitable topologies for specific optimization tasks. Ultimately, this contributes to making swarm-based optimization more transparent, robust, and trustworthy.
\end{abstract}

\ccsdesc[300]{Computing methodologies~Search methodologies}
\ccsdesc[500]{Computing methodologies~Metaheuristics}
\ccsdesc[500]{Computing methodologies~Explainable AI}



\maketitle
\section{Introduction} \label{sec:Intro}

The efficacy of PSO~\cite{kennedy1995new} is significantly influenced by its communication topology—the schematic arrangement that dictates how individual solutions, termed 'particles', share information. This architecture is critical as it directly impacts the swarm's ability to efficiently explore and exploit the search space, thus affecting overall algorithm performance~\cite{liu2016topology}. Traditional topologies such as Ring, Star, and Von Neumann offer distinct dynamics that can either hinder or enhance the swarm's optimization capabilities. Moreover, advancements in PSO have focused on augmenting its core strategies to overcome challenges such as premature convergence to local optima and enhancing its adaptability through methods like hybridization and strategic parameter adjustments~\cite{houssein2021major}.

Even with progress, PSO still runs into issues because it works like a "black box," leaving its decision-making unclear. This lack of clarity can make it hard to build trust and verify outcomes, especially in important areas like healthcare and self-driving systems. To tackle this problem, Explainable Artificial Intelligence (XAI)~\cite{dwivedi2023explainable} has come into play as a helpful way to make complicated algorithms easier to understand and more accountable. 
In this work, we aim to enhance the explainability and trustworthiness of PSO by exploring its dynamics under different communication topologies. Our objectives are to
\begin{enumerate}
    \item    Analyze how various topologies influence the search efficacy and convergence behaviors of PSO.
    \item 	Employ XAI methodologies to provide insights into the decision-making processes of PSO, enhancing the explainability and trustworthiness of the optimization results.
    \item 	Evaluate the performance of PSO under each topology using benchmark functions to determine optimal configurations for different types of optimization problems.
\end{enumerate}
 
 \section{Methodology} 

Explainable Artificial Intelligence (XAI) is becoming increasingly crucial in AI and machine learning. It enhances the transparency, trustworthiness, and interpretability of AI models, which is essential as these models progressively affect more areas of everyday life and critical decision-making. It helps clarify how these algorithms process and behave during search operations, which often involve intricate interactions and adjustments~\cite{chandramouli2023interactive}. Utilizing XAI techniques enables both researchers and practitioners to uncover how various algorithm components affect overall performance, observe how solutions develop over time, and understand the factors that drive algorithms towards optimal solutions.

\subsection{Particle Swarm Optimization (PSO)}

PSO~\cite{kennedy1995new} optimizes an objective function by refining candidate solutions iteratively. Each "particle" represents a potential solution, and its position is updated based on its experience and its neighbors' positions.

\begin{equation}\label{eq1}
  v_i (t+1)=wv_i (t)+c_1r_1(p_{best,i}-x_i (t))+c_2r_2(g_{best}-x_i (t))     
\end{equation}
\begin{equation}\label{eq2}   
x_i (t+1)=x_i (t)+v_i (t+1)    
\end{equation}


Equations \ref{eq1} and \ref{eq2} are the velocity and position update in PSO. Here, $v_i (t)$ and $x_i(t)$ are the velocity and current position of particle $i$ at time $t$, other parameters having their standard definitions as mentioned in Table~\ref{tab2}.


In this research, we consider three topologies: Ring, Star, and Von-Neumann, and analyze their configurations for better explainability.
In the Star topology, each particle is influenced by the best known position globally~\cite{miranda2008stochastic, ni2013new}. In Ring topology, each particle shares information with its  $k$ nearest neighbors, encouraging diversity and exploration~\cite{liu2016topology,sun2023particle}. In Von Neumann topology, particles are connected in a grid, with communication based on Delannoy numbers, promoting structure~\cite{von1935complete, lynn2018population}.

\subsection{Explainer Framework for Swarm Intelligence}

In this work, we adapted the IOHxplainer framework to analyze PSO behavior across hybrid topologies, enhancing it with XAI methods to assess the impact of different configurations. The framework supports continuous, integer, and categorical parameters—similar to SMAC~\cite{garcia2023explainable, lindauer2019boah}—and accounts for hyper-parameter dependencies for detailed PSO configuration.

Experiments use either grid or stochastic sampling methods, with an evaluation budget for each PSO configuration tested against the Black-Box Optimization Benchmark (BBOB)~\cite{hansen2009real}, which includes 24 noiseless functions in the COCO framework~\cite{hansen2022anytime}. SHAP-based XAI methods~\cite{lundberg2020local} are used to determine the marginal impact of each parameter on PSO’s performance, aiding in the interpretability of the results.

This is usually done with a swarm plot, giving PSO designers and users a clear picture of how different hyperparameters affect performance in various situations and across different setups.

\begin{figure*}[h] 
    \centering
    \begin{minipage}{0.7\textwidth}
        \includegraphics[width=\textwidth]{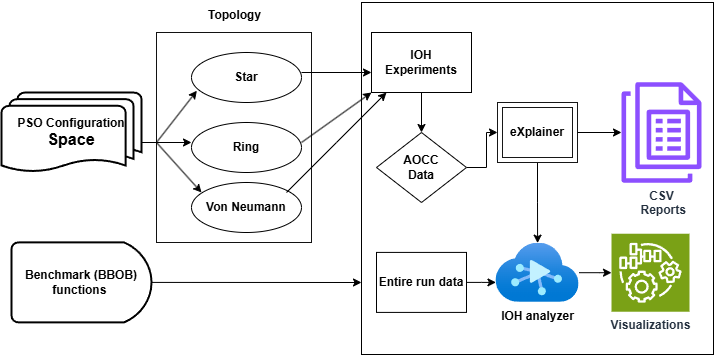}  
    \end{minipage}\caption{The proposed framework for PSO configuration for automate and explainable analysis.}\label{fig1}
\end{figure*}

To use the IOHxplainer in swarm intelligence studies, we start with a phase dedicated to gathering data, which might involve large grids or random samples of configurations. You can find a more detailed explanation of the framework in Figure~\ref{fig1}. The results are summarized in the following sections and can also be found on this GitHub link: \textcolor{blue}{\url{https://github.com/GitNitin02/ioh_pso}}.



\section{Experiment Setups and Result Discussion}
This section details the experimental setup and configuration of PSO and its associated topologies, focusing on its explainability through the explainable framework. For comprehensive evaluation, we apply this framework to a diverse set of benchmark functions.\\



\begin{table}[h]
    \centering
    \caption{PSO module and their configurable hyperparameter for each topologies for d=2.}\label{tab2}
    \begin{tabular}{|l|c|c|}
        \hline
        \textbf{Hyperparameter} & \textbf{Shorthand} & \textbf{Domain} \\
        \hline
        Cognitive coefficient & $c_1$ & \{0.3, 0.5, 0.7, 0.9\} \\
        Social coefficient & $c_2$ & \{0.2, 0.4, 0.6, 0.7\} \\
        Inertia weight & $w$ & \{0.9, 1.2, 0.4\} \\
        Number of particles & $n$ & \{50, 100, 150\} \\
        Nearest $k$ neighbors & $k$ & \{1, 2, 3\} \\
        Minkowski $p$-norm & $p$ & \{1, 2\} \\
        Delannoy numbers & $r$ & \{1, 2\} \\
        \hline
    \end{tabular}
\end{table}
The BBOB functions include a combination of unimodal, multimodal, and highly multimodal functions. Functions f1, f2, f5 to f14 are unimodal, f3, f4, f15, f16, f20, f23, and f24 are multimodal, and f17, f18, f19, f21, and f22 are highly multimodal.

We obtain results on all 24 BBOB functions in 2 d for every configuration we examine, with a budget of 100 iterations in Star, Ring and Von Neumann topologies. Table~\ref{tab2} presents the basic configuration of PSO with respect to each topology for 2 d. Each function's initial five instances are used, and each instance is run five times independently. The normalized Area Over the Convergence Curve (AOCC~\ref{eq3}) is a performance metric used to evaluate the efficiency of optimization algorithms and is defined as:

\begin{equation}\label{eq3}
    AOCC(\bar y) = \frac{1}{B} 
\sum_{i=1}^{B}
 \left(1- \frac{min(max((yi),lb),ub)-lb}{ub-lb}\right) 
\end{equation} 

Here, $\bar y$ is the series of optimal function values found so far, $B$ is the budget, and $lb$ and $ub$ are the lower and upper bounds of the function value range. Empirical cumulative distribution function (ECDF) is a statistical tool used for distribution of data. AOCC is the area under the ECDF curve with infinite targets inside the limits. Prior to computing AOCC, function values are log-scaled in accordance with BBOB rules~\cite{hansen2009real}, with limitations of -5 and 5 for the 2 d functions. These experiments are conducted using standard computational resources: Intel i7 CPU, 32 GB RAM, 8 cores. 

\section{Contribution of Hyperparameters} We examine 1,728 PSO configurations across 24 BBOB functions, focusing on the impact of different communication topologies and hyperparameters on performance. Using SHAP-based plots for 2 d (Table~\ref{tab2}), we analyze the contribution parameters such as n\_particles, $c_1$, $c_2$, $p$ etc. to the optimization. The SHAP values indicate how each parameter affects the AOCC, with yellow dots representing positive contributions and violet dots indicating negative ones. Notably, a lower $c_1$ value consistently improves performance, while a larger $c_1$ worsens it. For unimodal functions, a smaller n\_particle(50) performs best in a ring topology, while for multimodal functions, a larger value (100) works better. In contrast, the star topology benefits from a smaller n\_particle for highly multimodal functions.

\begin{figure*}[h]
    \centering
    \begin{minipage}{0.40\textwidth}
        \includegraphics[width=\linewidth]{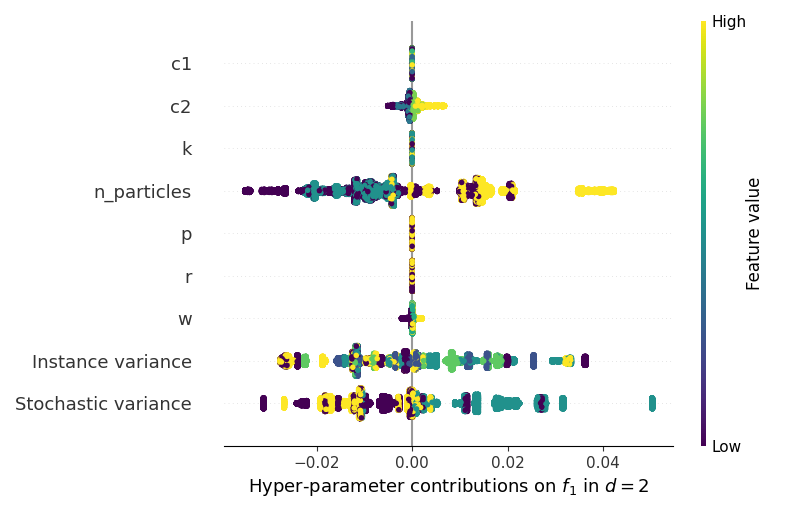}
    \end{minipage}
    \hfill 
    \begin{minipage}{0.29\textwidth}
        \includegraphics[width=\linewidth]{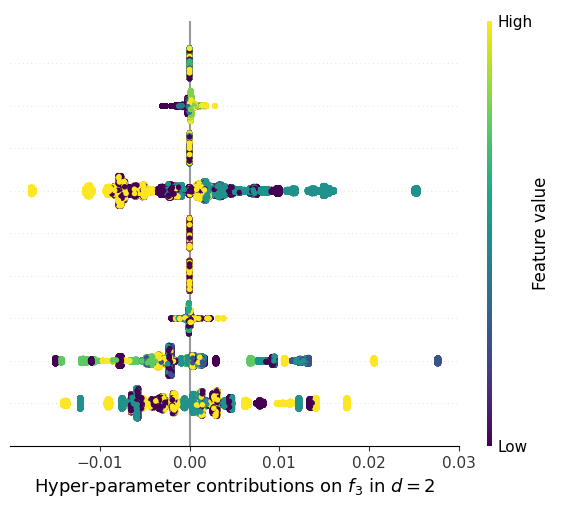}
    \end{minipage}
    \hfill
    \begin{minipage}{0.29\textwidth}
        \includegraphics[width=\linewidth]{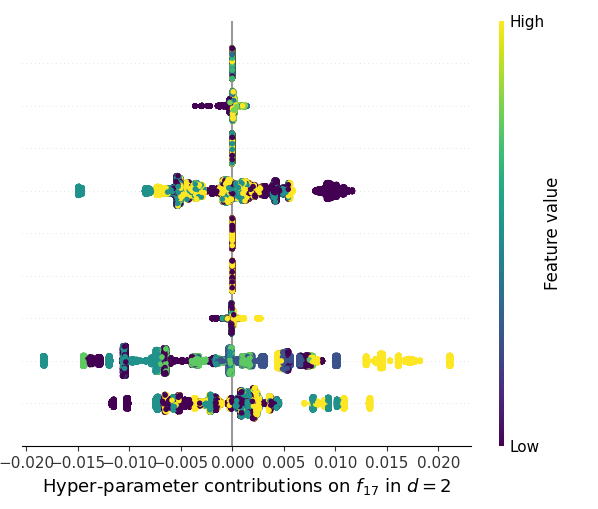}
    \end{minipage}
  \caption{Hyper-parameter contributions per benchmark function for d=2 for PSO using Star Topology.} \label{fig2}
\end{figure*}

Higher SHAP values correlate with better accuracy. In Figure \ref{fig2}, for the Star topology, the unimodal function (f1) shows yellow dots on the positive side of n\_particles, indicating improved accuracy. The multi-modal function (f17) displays blue dots, suggesting that higher values reduce accuracy. Green dots on the positive side of the multimodal function (f3) indicate a moderate SHAP value with a modest accuracy impact.
\subsection{Algorithm/configuration selection}
In this experiment, we computed the ``single-best'' and ``avg-best'' configurations for each topology---Star, Ring, and Von Neumann. We identified the best configuration per BBOB function (``single-best mean'') and the best overall (``avg-best mean''), comparing them with average performance and standard deviations. As shown in Table~\ref{T2}, PSO performs well on unimodal and simpler multimodal functions, showing low variance and stable convergence. However, it struggles on complex multimodal functions, particularly under the Star topology, due to poor escape from local optima. Functions like f3, f5, and f9 show strong performance. The Ring topology improves exploration through localized information sharing but converges slower and remains challenged on difficult landscapes. Full results are provided in the repository made available on  \textcolor{blue}{\url{https://github.com/GitNitin02/ioh_pso}}.

Due to space constraints, refer to the provided GitHub link for detailed experiment results. Based on Table~\ref{T2} settings, the Ring topology outperforms the Von Neumann and Star topologies in terms of the $R^2$ value, which typically indicates a better fit for regression tasks. While the Star topology shows a higher mean, the Ring topology is the superior choice for regression tasks focused on maximizing the $R^2$  value. Our analysis shows that the Von Neumann topology is the most time-efficient for PSO, taking 6.56 hours to run all 24 functions at dimension 2. The Star topology takes 7.2 hours, and the Ring topology takes 9.35 hours. Von Neumann’s efficiency is due to its structured communication, while the Ring topology’s higher time complexity is likely due to its sequential neighbor-based exchange. This highlights the impact of topology on PSO's computational cost and efficiency.

\begin{table*}[t]\centering
 \caption{Performance of Star, Ring and Von Neumann Topologies over all BBOB functions.}
\label{T2}
    \centering
   \scriptsize \begin{tabular}{|c|c|c|c|c|c|c|c|c|}
    \hline
      \textbf{Function}  & \textbf{Topologies} &  \textbf{single-best mean}  & \textbf{single-best std} & \textbf{avg-best mean} & \textbf{avg-best std} & \textbf{all mean} & \textbf{all std}  & \textbf{R2 train}\\ \hline
      
        \multirow{3}{*}{f1} & Star &	2.50E-01	&	5.40E-02	&	2.23E-01	&	2.62E-02	&\textbf{\color{red}{ 2.34E-01}}		&	4.18E-02& 9.76E-01
\\ 
        ~ & Ring &2.50E-01	&	5.40E-02	&	2.23E-01	&	2.62E-02	&	2.32E-01	&	4.32E-02  & \textbf{\color{red}{9.83E-01}}

   \\ 
        ~ & Von Neumann &2.50E-01	&5.40E-02	&2.23E-01	&2.62E-02	&2.33E-01	&4.25E-02
 &9.80E-01
  \\ \hline

        \multirow{3}{*}{f2} & Star&	2.35E-02	&	2.32E-02	&	1.70E-02	&	1.89E-02	&\textbf{\color{red}{ 1.86E-02}}		&	2.23E-02& 9.86E-01
\\ 
        ~ & Ring &	2.39E-02	&	2.67E-02	&	1.70E-02	&	1.89E-02	&	1.84E-02	&	2.23E-02 & 9.87E-01
   \\ 
        ~ & Von Neumann & 2.39E-02	&2.67E-02	&1.70E-02	&1.89E-02	&1.85E-02	&2.22E-02
 & \textbf{\color{red}{9.88E-01}}
 \\ \hline

        \multirow{3}{*}{f3} & Star&	1.01E-01	&	2.82E-02	&	1.01E-01	&	2.82E-02	&\textbf{\color{red}{ 9.32E-02}}		&	2.09E-02& 9.66E-01
\\ 
        ~ & Ring &	1.01E-01	&	2.82E-02	&	1.01E-01	&	2.82E-02	&	9.27E-02	&	2.12E-02 &\textbf{\color{red}{9.77E-01}} 
   \\ 
        ~ & Von Neumann & 1.01E-01&	2.82E-02	&1.01E-01	&2.82E-02	&9.30E-02	&2.11E-02
 &9.69E-01
  \\ \hline

        \multirow{3}{*}{f4} & Star	&	9.62E-02	&	2.99E-02	&	7.81E-02	&	1.45E-02	&\textbf{\color{red}{ 8.68E-02}}		&	2.23E-02& 9.87E-01
\\ 
        ~ & Ring &	9.51E-02	&	3.13E-02	&	7.81E-02	&	1.45E-02	&	8.62E-02	&	2.26E-02 &\textbf{\color{red}{9.93E-01}} 
  \\ 
        ~ & Von Neumann & 9.87E-02&	3.06E-02&	7.81E-02&	1.45E-02&	8.66E-02	&2.24E-02
 &9.83E-01
  \\ \hline

        \multirow{3}{*}{f5} & Star&	1.55E-01	&	2.72E-02	&	1.55E-01	&	2.72E-02	&	1.37E-01	&	2.33E-02&\textbf{\color{red}{ 9.96E-01}} 
\\ 
        ~ & Ring &	1.55E-01	&	2.72E-02	&	1.55E-01	&	2.72E-02	&\textbf{\color{red}{ 1.38E-01}}		&	2.26E-02 & 9.94E-01
   \\ 
        ~ & Von Neumann & 1.55E-01&	2.72E-02&	1.55E-01&	2.72E-02&	1.37E-01	&2.32E-02
 &9.90E-01
  \\ \hline

        \multirow{3}{*}{f6} & Star&	1.41E-01	&	3.49E-02	&	1.38E-01	&	2.33E-02	&\textbf{\color{red}{ 1.33E-01}}	&	2.85E-02& 9.79E-01
\\ 
        ~ & Ring &	1.38E-01	&	2.33E-02	&	1.38E-01	&	2.33E-02	&	1.31E-01	&	2.92E-02 &\textbf{\color{red}{ 9.94E-01}} 
   \\ 
        ~ & Von Neumann & 1.44E-01 &	3.46E-02&	1.38E-01	&2.33E-02	&1.32E-01&	2.89E-02
 &9.78E-01
  \\ \hline

        \multirow{3}{*}{f7} & Star&	1.92E-01	&	3.16E-02	&	1.78E-01	&	4.26E-02	&\textbf{\color{red}{ 1.84E-01}} 		&	3.81E-02 & 9.64E-01
\\ 
        ~ & Ring &	1.91E-01	&	4.60E-02	&	1.78E-01	&	4.26E-02	&	1.83E-01	&	3.81E-02 &9.80E-01
   \\ 
        ~ & Von Neumann &1.91E-01&	4.60E-02&	1.78E-01&	4.26E-02	&1.83E-01	&3.80E-02
 & \textbf{\color{red}{ 9.82E-01}}
  \\ \hline

        \multirow{3}{*}{f8} & Star&	1.46E-01	&	4.30E-02	&	1.33E-01	&	5.85E-02	&	1.34E-01	&	4.82E-02& 9.69E-01
\\ 
        ~ & Ring &	1.42E-01	&	4.32E-02	&	1.33E-01	&	5.85E-02	&	1.34E-01	&	4.80E-02 &\textbf{\color{red}{ 9.84E-01}} 
  \\ 
        ~ & Von Neumann & 1.44E-01 &	3.82E-02	&1.33E-01&	5.85E-02&	1.34E-01&	4.82E-02
 &9.77E-01
  \\ \hline

        \multirow{3}{*}{f9} & Star&	1.63E-01	&	3.81E-02	&	1.63E-01	&	3.81E-02	&	1.43E-01	&	4.74E-02& 9.54E-01
\\ 
        ~ & Ring &	1.63E-01	&	3.81E-02	&	1.63E-01	&	3.81E-02	&	1.42E-01	&	4.74E-02 &\textbf{\color{red}{ 9.88E-01}} 
   \\ 
        ~ & Von Neumann & 1.63E-01&	3.81E-02	&1.63E-01&	3.81E-02&	1.43E-01&	4.88E-02
 &9.63E-01
  \\ \hline

        \multirow{3}{*}{f10} & Star&	2.36E-02	&	3.04E-02	&	2.36E-02	&	3.04E-02	&	\textbf{\color{red}{ 1.60E-02}} 	&	1.85E-02& 9.61E-01
\\ 
        ~ & Ring &	2.36E-02	&	3.04E-02	&	2.36E-02	&	3.04E-02	&	1.54E-02	&	1.83E-02 & \textbf{\color{red}{ 9.82E-01}}
   \\ 
        ~ & Von Neumann &2.36E-02&	3.04E-02&	2.36E-02&	3.04E-02&	1.57E-02&	1.85E-02
 &9.54E-01
  \\ \hline

        \multirow{3}{*}{f11} & Star &	6.00E-02	&	4.22E-02	&	6.00E-02	&	4.22E-02	&	6.00E-02	&	4.01E-02  &  9.80E-01
\\ 
        ~ & Ring &	6.00E-02	&	4.22E-02	&	6.00E-02	&	4.22E-02	&	6.00E-02	&	4.01E-02  & 9.80E-01
   \\ 
        ~ & Von Neumann &5.90E-02	&4.22E-02&	6.00E-02&	4.22E-02&	6.00E-02	&4.01E-02
 &9.77E-01
  \\ \hline

        \multirow{3}{*}{f12} & Star 	&	1.73E-02	&	1.78E-02	&	1.53E-02	&	1.28E-02	&	1.35E-02	&	1.31E-02 & 9.43E-01

\\ 
        ~ & Ring &	1.62E-02	&	1.30E-02	&	1.44E-02	&	1.25E-02	&	1.35E-02	&	1.26E-02  &\textbf{\color{red}{ 9.91E-01}} 
   \\ 
        ~ & Von Neumann & 1.84E-02 &	1.53E-02&	1.84E-02&	1.53E-02&	\textbf{\color{red}{ 1.36E-01}} &	1.28E-02
 &9.58E-01
  \\ \hline

    \end{tabular}
\end{table*}

\section{Conclusion:}
This research advances the explainability of PSO by investigating how different communication topologies affect its performance. Using a modified IOHxplainer framework, we evaluated 24 benchmark functions with multiple instances and seeds. Our findings reveal that Von Neumann topology enhances diversity and gradual convergence, making it suitable for complex optimization landscapes. Ring topology offers a balanced trade-off between exploration and exploitation, while Star accelerates convergence but may cause early stagnation.
Future directions include dynamic hyperparameter tuning, scaling to higher dimensions, and applying the framework to evolving neural networks and high-stakes domains like cybersecurity. Additionally, we plan to extend the IOHxplainer framework to study evolving neural networks, focusing on reducing their structural complexity while maintaining performance.

\section*{Data availability:} \textcolor{blue}{\url{https://github.com/GitNitin02/ioh_pso}}

\section*{Acknowledgment}
\small This work was supported by Dr B. R. Ambedkar National Institute of Technology Jalandhar, the Anusandhan National Research Foundation, Government of India (Award No. MTR/2021/000503), the Australian Researcher Cooperation Hub through the Australia-India Women Researchers' Exchange Program, and the Spanish Ministry of Economy and Competitiveness through the Ramón y Cajal Research Grant (Award No. RYC2023-045020-I).
 
\bibliographystyle{ACM-Reference-Format}
\bibliography{acmart.bib}

\end{document}